# 基于海量在线历史数据与图卷积深度学习的大电网快速判稳方法

Haotian Cui　　Xianggen Liu,　　Yanhao Huang
Cht15@mails.tsinghua.edu.cn　　liuxg16@mails.tsinghua.edu.cn　　hyhao@epri.sgcc.com.cn
Tsinghua University, Beijing 100084, China
Department of Biomedical Engineering, School of Medicine,
IDG/McGovern Institute for Brain Research

Abstract：The paper intends to model the stability of power system with a deep learning algorithm to the problem, aiming to delay the removal of the fault. The so-called "fail-delay cut-off" refers to the occurrence of N-1 backup protection action on the backbone network of the system, resulting in longer time for the removal of the fault. In practice, through the analysis and calculation of a large number of online data, we have found that the N-1 failure system of the main protection action will not be unstable, which is also a guarantee of the operation mode arrangement. In the case of the N-1 backup protection action, there is an approximately 2.5% probability that the system will be destabilized. Therefore, research is needed to improve the operating arrangement.

## 引言：

电力系统运行过程中的快速稳定判断, 对系统的稳定运行, 风险预警起到了至关重要的作用。基于海量历史数据的快速判稳，是数据挖掘方法在电力系统中重要的应用场景。
目前，该领域国内外的相关研究大致按照一下技术路线：1）对电力系统仿真计算数据进行清理；2）根据已有的电力系统知识选取表征系统运行特性和稳定性的特征量；3）对初筛的特征量进行挑选和压缩；4）选用适当的机器学习算法进行判稳模型训练和测试。研究重点主要集中在特征量筛选和学习算法改进方面。
面向实际大系统，这些成果主要存在以下不足：
（1)）选取的特征量通常包括静态和动态物理量，后者需要进行暂稳计算。为了与实际应用相配合，暂稳计算一般设计成仅进行几步迭代以减少耗时，但依旧需要一定的硬件资源，工程实现也较复杂； 2）工作的基础一般是小型(算例)系统，缺乏针对实际大系统海量在线历史数据特点的模型和方法研究；

另一方面，近年来，以深度卷积网络为代表的深度学习[LeCun Y, Bengio Y, Hinton G. Deep learning[J]. nature, 2015, 521(7553): 436.]在数据挖掘和人工领域取得了长足发展。例如 2006 年 Geoffrey Hinton 提出深度置信网（Deep Belief Net：DBN）[2]，Ruslan Salakhutdinov 提出的深度波尔兹曼机（Deep Boltzmann Machine :DBM）[34]。2012 年，Hinton 在 ImageNet 竞赛上运用卷积神经网络（Convolutional Neural Networks）取得突破，首次提出卷积操作在图像数据上提取特征的重要作用[49]。2015 年微软何凯明团队利用 152 层网络在 ImageNet 比赛上将错误率降低到 3.57% [7]，2016 年，以对抗生成网络[48]（Generative Adversarial Networks）为代表的图片生成算法提出一个生成器和一个判别器相互竞争的学

习方式，为深度学习提供了一种新的训练方法，得到广泛运用[15]。深度学习相对于传统方法，可以实现端到端学习，避免了人工挖掘特征的偏差；模型复杂度高，具有高度的表达能力，能够适应在如电力系统的海量大数据中学习复杂特征，挖掘出较传统数据驱动和特征工程方法更深入的特征。此外，电力网络中的母线和交流线，可以看作节点和连边，构成高维的图结构，对于图结构进行优化的深度学习神经网络，有潜力对电力网络更好地建模。

常见的深度学习模型包括深层全连接网络，卷积神经网络，循环神经网络和堆栈自编码器等。由于深度学习的迅速发展，将相关方法用于电力领域有见报道【基于深度置信网络的电力系统暂态稳定评估方法】。但深度学习模型尤其适用的问题特征，如卷积神经网络适合于处理二维欧式空间上的图像数据，或三维欧式空间的视频数据；又如循环神经网络适合处理语言，音频等时序信号。而电力网络元件之间互有连接，组成高维图结构，不适合上述常见深度学习模型处理。

针对上述问题，本文首次将图卷积网络应用在电力系统问题中，提出一种以图卷积神经网络为主的，结合了深层全连接网络，卷积神经网络等多种深度学习模型的电力系统暂态稳定评估方法。设计流程为 “记录-训练-应用”，利用当天潮流计算数据更新训练模型，用于次日的电网判稳。与其他相关文章比较[2，3]主要应用在新英格兰 39 节点示例数据，而本文在海量真实在线数据中进行了实验分析，具有真实应用价值；主要应用常见的深度学习模型，而本文利用首次图卷积深度学习，首次直接利用了电网的图结构特征。

学习满足上述数据本文运用实际在线历史数据，将深度学习应用在电力系统网络快速稳定判断中。所研究的系统计算节点达到 6000 个。考虑到电网运行的实际情况，主要研究在线计算的失稳故障预判问题。具体研究为某区域 500kv 线路的后备保护动作。

本文以实际应用为导向，重点在海量在线计算历史数据的数据特征挖掘，失稳样本比例失衡的处理，适用于电力网络的深度学习网络结构，电力网络全局与局部特征的分析与表达方面进行了研究。
本文的主要贡献有以下几点：

1. 利用深度学习网络实现了电力网络在线数据端到端的自动学习，并借鉴自然语言领域将电力网络节点向量化，有效发掘出可学习的网络特征，实现了准确的在线失稳故障集预测。
2. 发现全局特征和局部特征的结合，有助于网络状态的建模和失稳故障的判断。并选取了有效的全局和局部特征，利用深度学习进行有效建模；
3. 创新地将局部特征表达为高维图结构，将改进后的图卷积神经网络应用在电力网络中，提出了高度适用于电力网络问题的图卷积神经网络结构。

# 问题定义：

这一段请黄老师帮助修改
对于实际大系统在线稳定分析，其所积累的历史数据体量极其庞大，且随时间推移还在不断生成。通常，大系统在线分析每 15min 计算周期产生的数据就有 80M，全天可达 8GB，全年可达 2.9 PB。 若直接全部使用这些数据进行快速判稳，在现有的软硬件条件下，由于处理能力有限将难以获得很好的结果，因此必须考虑对其进行有效的信息浓缩。 同时，还应从实际大系统运行特性出发，研究数据规律，寻找解决问题的快捷途径，并与在线计算的需

求相协调。
利用断面数据，预判特定故障是否会导致电力系统失稳，为本文的研究问题。利用机器学习方法，从静态潮流断面中进行快速判稳，能够有效节省线上计算压力。本文采用某电网 17 年 1 月真实线上潮流数据，构造故障集。共涉及一周内的潮流断面 1440 个，故障总样本数 35 万。故障集经仿真判断，共有系统失稳样本 3.5 万个，占比 10%。

# 深度卷积网络（convolutional neural net，CNN）

卷积网络是一种从大脑视觉皮层借鉴得到的特殊神经网络。主要利用可学习的卷积核在数据上做卷积操作，学习得到相关数据特征，在图像识别，分割，视频分析领域近年来有优异表现。卷积网络主要由卷积层（特征提取层），池化层（采样层）组成。
卷积层的卷积核在输入数据上进行操作，达到抽取特征的效果，如式（1）所示：

$$m^{(l+1,j)} = f(m^{(l)} * k_j^{(l)} + b_j^{(l)}) \qquad (1)$$

其中 $m^{(l+1,j)}$ 为第 l 层卷积层输出的的第 j 个特征图（feature map）；$m^{(l,j)}$ 为 l 层卷积层的特征图输出；$k_j^{(l)}$ 为第 j 个卷积核；$b_j^{(l)}$ 为对于第 j 个卷积图的偏置向量。

池化层从卷积层输出的特征图中进行最大值采样，获取重要特征并降低特征图的维度。

$$out(N_i, C_j, h, w) = \max_{m=0}^{kH-1} \max_{n=0}^{kW-1} input(N_i, C_j, stride[0] * h + m, stride[1] * w + n)$$

CNN 主要用来识别位移、缩放及其他形式扭曲不变性的二维图形。由于 CNN 的特征检测层通过训练数据进行学习，所以在使用 CNN 时，避免了显示的特征抽取，而隐式地从训练数据中进行学习；且由于同一特征映射面上的神经元权值相同，所以网络可以并行学习，这也是卷积网络相对于神经元彼此相连网络的一大优势。卷积神经网络以其局部权值共享的特殊结构在图像处理方面有着独特的优越性，其布局更接近于实际的生物神经网络，权值共享降低了网络的复杂性，特别是多维输入向量的图像可以直接输入网络这一特点避免了特征提取和分类过程中数据重建的复杂度。

# 图卷积网络（GCN）

图卷积网络是自 2010 年由 lecun 等[]首先提出，用于处理非欧式结构的数据，对图结构数据有比卷积网络更强的表达能力，经过几年的发展出了多个分支，如运用谱分析理论进行图卷积等[Bruna et al., ICLR 2014; Henaff et al., 2015]，特别是 17 年 Kipf 等提出的图卷积网络（graph convolutional network）[]。用图卷积网络表达和刻画电网能够准确抓住电网中潮流传播的特征，相较于上述方法更符合电网的物理特性。

对于图数据 G=（$\nu$，$\varepsilon$），图卷积网络是做用在图结构的数据上的卷积网络，将共享的卷积核作用在整个图上，拟合根据图结构上数据的特定目标函数。具体上一层图卷积网络需

要两部分输入，第一是对图上每个节点 i 的特征描述 $x_i$ ，用维度为 $N \times D$ 的矩阵 $H^{(1)}$ 表示（N 为图上节点个数，D 为每个节点的描述特征维度）；第二是用于描述图结构的矩阵，这里使用图的邻接矩阵 A。经图卷积处理之后，输出矩阵 $H^{(l+1)}$ 。则整个图卷积网络由多层图卷积操作层构成，对于第 l 层的最简图卷积操作可由下式表达：

$$f(H^{(l+1)}, A) = \delta(AH^{(l)}W^{(l)})$$

这里 $W^{(l)}$ 是第 l 层的权重矩阵，σ为该层 d 的非线性激活函数，这里选择 ReLU[]。本简化图卷积操作已经可以完成有效建模。在此基础上增加两点改进，第一由于邻接矩阵的对角线元素为 0，做改进将单位阵与邻接矩阵 A 相加，引入对节点自身特征的考虑。第二由于在多层卷积之间会多次与 A 做矩阵乘法，会使得输出的特征向量大小改变导致影响网络稳定性，将邻接矩阵 A 归一化使得每行的和是 1，以避免上述问题。得到

$$f(H^{(l)}, A) = \delta(D^{-\frac{1}{2}}AD^{-\frac{1}{2}}H^{(l)}W^{(l)})$$

## 1.3 学习方法

本文所用上述模型均采用交叉熵为训练优化目标：

$$L = -\frac{1}{N}\sum_{i=1}^{N} y_i \log_2(\dot{y}_i)$$

式中：$y_i$ 为训练集第 i 个样本的实际值（系统失稳/系统稳定）；$\hat{y}_i$ 为对应的模型预测值。上述模型均通过梯度反向传播[]方法进行参数训练，具体使用 Adam[]优化策略，学习速率均设置为 0.001.

# 2 暂态稳定模型

## 2.2 电力网络图定义

请黄老师补充
电力网络是一种典型的复杂网络，用电力系统的电器结线图可直观地映射成图卷积所要的图数据，之后利用图卷积网络进行建模。在电器系统中包含变压器，发电机，母线，交流线等设备；这里将母线视为图中节点，将其他设备视作母线之间的连边。
1.3 有什么参考文献这样做吗
1.4 定义的优势
这样的定义还原了电力网络的结构，同时方便神经网络挖掘其中的特征；一方面对于网络中的故障节点可以在途中清晰的反映出局部拓扑关系，按照电器物理连接，故障的冲击在电网

中的传播模式与物理连接关系有高度相关性，图结构可以同时保留潮流状态参数和链接拓扑信息，使得深度网络等机器学习方法学习出电网传播模式的相关性。
由此定义对于潮流断面我们可以得到如下网络结构表示：
状态矩阵 S 和邻接矩阵 A；

# 特征量选取

本文利用潮流断面的静态物理量进行稳态计算，因此可以在不用暂态稳定计算的前提下进行预测。在这种情况下，首先探究系统的稳定状态和哪些系统信息有关。一般情况下，随着电力系统运行状态的变化，系统的稳定状态随之变化。本文认为电力系统的稳定性，特别是 N-1 故障发生时系统稳定性与两大方面信息相关，系统的全局状态和故障的局部特性。例如，当系统整体稳定时，某危险故障可能并未导致网络失稳；同样的故障在系统处于非稳定的临界状态时反而导致系统整体失稳。因此我们将进行稳定性判断需要的特征分为表征电力网络的全局特征和故障点局部特征两部分。

具体包括

a) 全局特征量包括

本文对全局特征量的描述主要包括了两种形式：
第一种是将整个电力网络的原始状态量直接作为特征量，由深度卷积网络直接从网络中学习表征网络的特征量；用于描述的原始状态量集合为 S，S 包括如下表 1，2 中涉及的多种物理量。
第二种是文从电力网络的原始状态量抽取有效的统计特征，用于描述系统全局状态的统计特征量可由式（1）表示：

$$F_i = f(p_i, s_i, r_i) \qquad (1)$$

式中：: Fi 为特征量；pi 为与该特征量对应的物理量， pi∈P，P 是用于构成特征量集的物理量集合，本文 中其仅包含静态物理量；si 为统计量，si∈S，S 是用 于构成特征量集的统计量集合；ri 为物理量的统计范围，ri∈R，R 是用于构成特征量集的统计范围集合
P 中的主要元素如表 1 所示。

| 物理量 | 含义 | 物理量 | 含义 |
| --- | --- | --- | --- |
| $V$ | 母线电压幅值 | $\cos\theta_L$ | 负荷功率因数 |
| $\theta$ | 母线电压角度 | $P_{AC}$ | 交流线路有功(取 I 侧) |
| $P_G$ | 发电机有功 | $Q_{AC}$ | 交流线路无功(取 I 侧) |
| $Q_G$ | 发电机无功 | $P_{DC}$ | 直流线路有功(取 I 侧) |
| $\cos\theta_G$ | 发电机功率因数 | $Q_{DC}$ | 直流线路无功(取 I 侧) |
| $P_L$ | 负荷有功 | $Q_{PC}$ | 并联电容器投入容量 |
| $Q_L$ | 负荷无功 | $Q_{PL}$ | 并联电抗器投入容量 |

图 1. 构成特征量的物理量

| 传统统计量 | 含义 | 稳健统计量 | 含义 |
| --- | --- | --- | --- |
| Max | 最大值 | $Median_L$ | 中位数 |
| Min | 最小值 | Msd | 中位数标准差 |
| Mean | 平均值 | 1st | 1/4 分位数 |
| Sd | 均值的标准差 | 3st | 3/4 分位数 |
| Skew | 偏度 | Mad | 中位绝对离差 |
| Kurt | 峰度 | Interq | 四分位差 |
| — | — | Mj10 | 10%截尾均值 |
| — | — | Mj10s | 10%截尾均值标准差 |

图 2. 构成特征量的统计量

按照上述策略，选出的特征量共有 556 个。

b) 局部特征量包括

当统计范围为故障点局部时，一种可能的假设是故障点周围一定范围内的节点，将在极短时限内受到故障冲击的影响，因此故障点周围的稳定状态将直接与故障冲击的传播相关。基于此，我们认为故障点局部特征可以作为故障网络稳定失稳判断的有效特征。

在电网实际运行数据中，我们选取故障发生时，我们从故障点的拓扑图中以故障母线为初始做广度优先搜索，选取临近 50 个节点组成的拓扑图作为故障点局部图。在此图中，选择表 2 所示的各个特征量作为输入特征。特别强调的是，我们认为对于电网的一个局部，应包含显性特征和隐性特征两部分：显性特征指可以在电网潮流断面的物理量中得到体现的特征，这部分特征表现了母线，交流线等被监测部分的电气状态；隐性特征则是由于如用户用电器，元件所在电气环境等会影响电气网络，而这部分因素并没有直接在潮流断面静态物理量中体现。本文认为，由于每个电器元件所处隐性因素各有不同，可以将隐性因素视为一种时不变的先验因素，本文创新性地借鉴自然语言处理领域词向量（word embedding 添加参考）的方法，为每一元件构建独有的可学习随机向量表征每个元件所在的隐性特征先验，通过深度网络学习训练阶段，得到对隐性先验的合理表征。总结地，为了能全面表现显性和隐性特则，本文将局部拓扑图的采取的特征分为两部分，第一部分是直接利用静态物理量来表征显性特征，第二部分是用可学习向量来表现隐性特征。

综上，我们选取的特征维度分两部分分别有：

(1) CNN 所使用的原始数据特征 G 为二维数据矩阵（7000, 13），包含全网 7000 条母线，每条母线有 13 维状态特征，共 91000 维数据点。
除原始数据特征外电网全局统计特征共 537 维

(2) 局部结构特征：局部结构特征首先由故障点为中心，按照电力网络原始的连接向外做广度优先搜索，取前 50 个节点构成，每个节点包含 59 维特征，组成局部的拓扑图结构（A, H）；矩阵 A 为局部 50 个图节点的邻接矩阵，维度（50, 50）.矩阵 H 为节点数据特征矩阵，维度（50, 59）。
除上述图特征外，另有 20 维的故障元件“词向量”特征。局部结构特征计 2970 维特征

# 深度网络模型与策略流程：

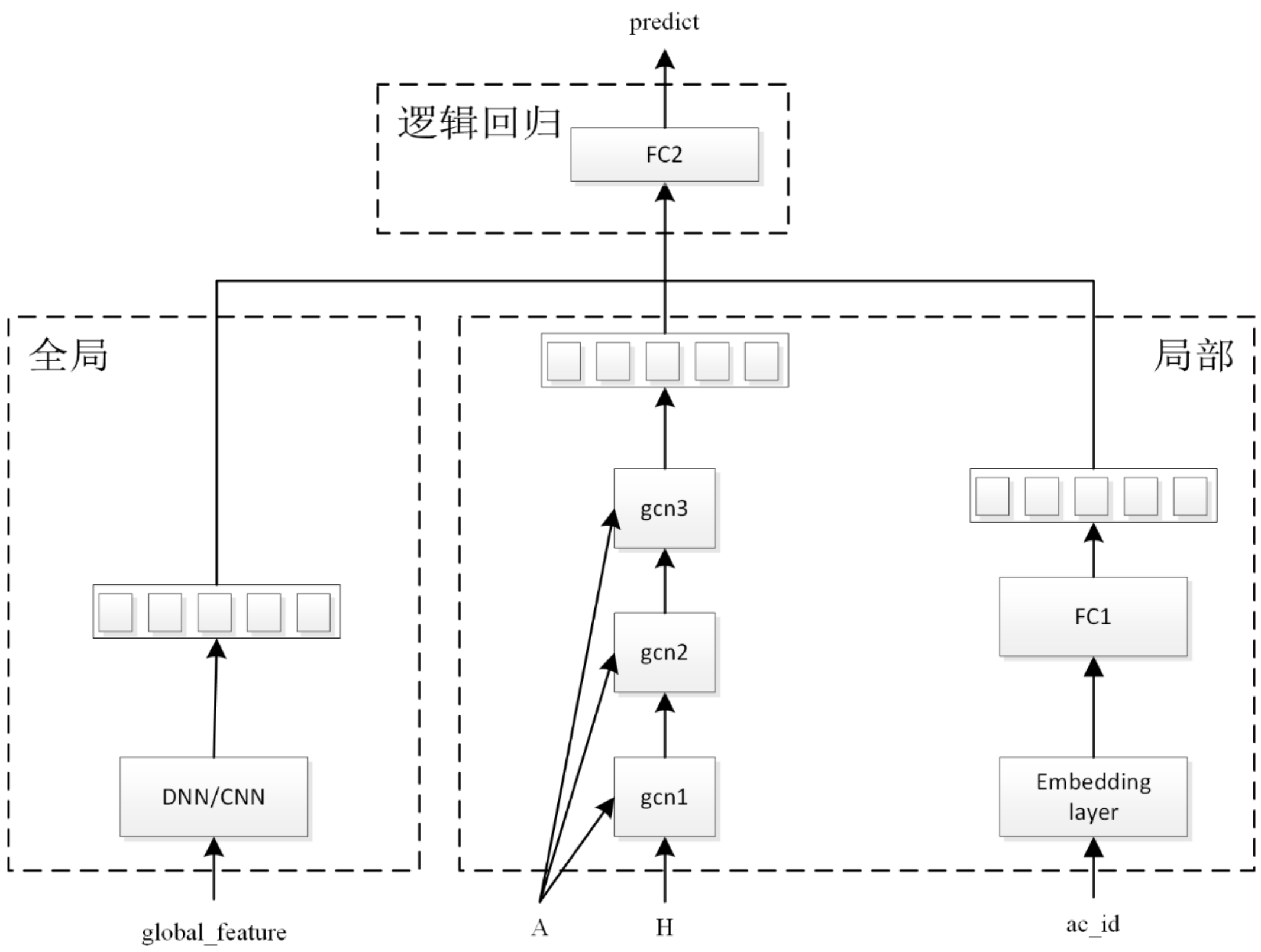

图 1.

如图 1.所示，模型分为全局网络，局部网络，逻辑回归分类器三部分：

1.全局网络部分，输入为选取的全局特征 G，用卷积网络或深层全连接网络编码为向量。选用卷积网络结构如下图所示：

经过神经网络得到全局特征向量 $S_g$

2.局部网络部分：

2.1 故障局部图结构特征包含上述提到的邻接矩阵 A 与局部特征矩阵 H，图卷积部分经过三层图卷积网络，输出得到局部特征向量 $S_l$

2.2 交流线元件向量化，用可训练的随机向量层（embedding layer）将故障交流线序号表示为 20 维的特征向量，再经过全连接神经网络得到故障向量特征 $S_{id}$

3. 逻辑回归器部分：

将上述三部分向量通过逻辑回归层得到最终预测失稳概率 y_predict

上述所用到的网络结构与参数量如下表：

# 高可靠性约束与策略流程

取 2016 年 1 月份实际大电网数据，对本文所关心区域 500 kV 线路后备保护动作故障进行了故障统计。在初始的计算结果中，系统失稳的情况占全部故障的 6.2%。 对于这样分布不均匀的样本数据，深度学习和机器学习算法都容易对“稳定”的故障产生过拟合。这种样本量巨大，但有意义样本比例很低的情况与大数据价值巨大但密度很低的特性相似。

针对此问题，首先考虑的是调整仿真数据以增加失稳样本的方法有两个困难；一是大电网仿真数据的调整策略不易确定，在保证潮流收敛的同时，又要保证获得的新增失稳样本对模型训练结果有实际影响比较困难；二是人为构造运行方式意味着样本准备的计算量将会显著增加，不易满足在线计算的时间要求。同时增加失稳样本也改变了数据的真实分布，裁剪冗余样本又降低了可学习样本量，综合考虑资源消耗、算法效果和实现难度，本文采用了两种方法来实现高可靠性约束：

一、降低判故障置信度阈值，达到可靠度要求。由于深度模型采用回归方法训练，整体是在拟合失稳的状态，越靠近 0 说明以越高的置信度判稳，当把阈值设置在合理的范围可以兼顾高可靠性和其他性能。

二、保证故障样本数量不变，以随机采样的方法，对故障节点数据集采样，使得采样得到的故障样本和稳定样本量一致。

本文结合图卷积深度网络的快速判稳模型训练的具体流程如图 4 所示。

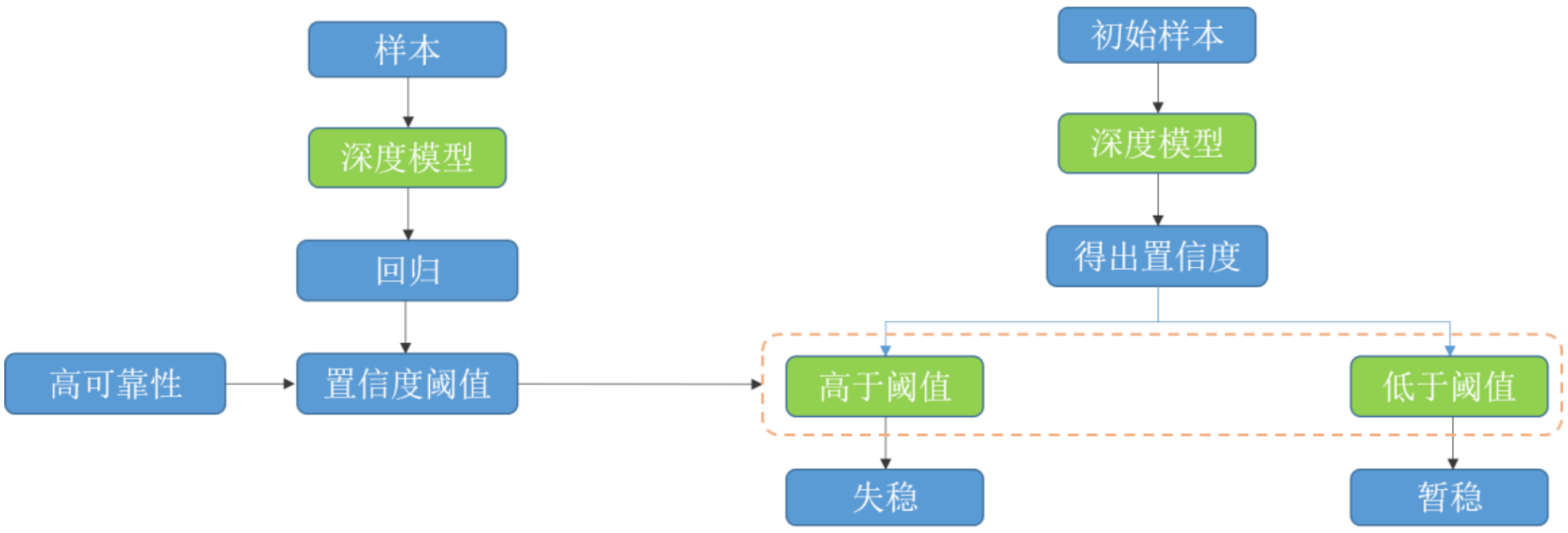

# 基准模型

1、 基准模型一

用前一天的标签来预测此时刻是否失稳。 一天大约有 200 个断面，将 200 个断面的标签做一个平均值. 以 20170106 为例，  为了达到 kkd 为 98%，  当均值大于 0.03 则为失稳。电网节点状态时序关联性较强，基于前一天预测结果的基准是很强的基准结果，有下表展示。

| 日期 | kkd | ryd | ysl | acc |
|---|---|---|---|---|
| 20170102 | 89.71 | 30.51 | 90.59 | 96.38 |
| 20170103 | 84.74 | 35.52 | 87.80 | 94.25 |
| 20170104 | 88.91 | 59.78 | 76.31 | 84.65 |
| 20170105 | 86.51 | 43.59 | 82.93 | 91.06 |
| 20170106 | 98.19 | 65.17 | 72.13 | 81.65 |
| 20170107 | 98.32 | 43.68 | 84.32 | 93.00 |

2、基准模型二

我们复现[1]中介绍的 SVM 方法，来判定电网的稳定状况，复现过程分为以下四步：1、扩展边界，以 2017/01/05 的所有断面为例：一共有 61601 个有效样本，其中有 7317 个失稳样本。 采用高罚值（C=20000, theta = 1/61601）的 SVM 做分类，共有 1149 个失稳样本，错判为稳定，有 141 个稳定样本错判为失稳，其中还有支持向量 3645 个。扩展后失稳样本量有 11103 个。2、压缩样本量：选取与分界面最近的 11103 个稳定样本，用作训练。

3、基准模型三

我们建立三层多层感知机来尝试解决此问题，隐层维度分别为 200 维、100 维，同样使用回归方法。

考虑到电网判稳的实际需求，我们采用如下四个指标：

1） 可靠度 K_sd:

$$K_{kd}=(S_{tf}-L)/S_{tf}\times 100\%$$

2） 准确率

3） 冗余度 R_yd:

$$R_{yd}=(P_{ds}-P_{df})/P_{ds}\times 100\%$$

4） 压缩率 Y_sl:

$$Y_{sl}=(S_f-P_{ds})/S_f\times 100\%$$

# 实际系统测试：

a) 展示在多天数据上的效果

SVM:

| 日期 | kkd | ryd | ysl | acc |
|---|---|---|---|---|
| 20170102 | 98.05 | 88.43 | 34.10 | 41.57 |
| 20170103 | 98.09 | 88.81 | 13.17 | 22.70 |
| 20170104 | 98.05 | 87.36 | 11.27 | 22.27 |
| 20170105 | 98.14 | 86.91 | 10.92 | 22.35 |
| 20170106 | 98.03 | 83.42 | 37.65 | 47.78 |
| 20170107 | 97.98 | 81.42 | 49.47 | 58.67 |

graphModel:

| 日期 | kkd | ryd | ysl | acc |
|---|---|---|---|---|
| 20170102 | 98.02 | 84.47 | 50.95 | 58.42 |
| 20170103 | 98.02 | 87.74 | 20.84 | 30.35 |
| 20170104 | 98.02 | 79.77 | 44.59 | 55.58 |
| 20170105 | 98.05 | 85.40 | 20.26 | 31.67 |
| 20170106 | 98.09 | 54.66 | 77.18 | 87.32 |
| 20170107 | 98.20 | 57.87 | 78.65 | 87.87 |

b) 展示 graph 效果与 mlp 对比，与 svm 对比，与 baseline 对比都要好；证明深度学习挖掘出了有效特征

用 2017 年 1 月 5 日数据作为训练数据集，一共 58293 个样本，对 2017 年 1 月 6 日的数据进行测试，一共 42942 个测试样本。下表展示了深度网络模型与三种基准模型在测试集上对于 4 个指标的表现情况，为了使各模型之间具有可比性，同时又满足电网判稳任务对于高可靠性的约束，我们通过调节阈值的方法，将可靠性稳定在 98%左右，再来观察在其他三个指标上的优劣，可以发现各个神经网络模型没有因为数据量的不均衡而出现偏差，同时观察到深度网络模型以很大的优势超过了其他基准模型，在三个指标上一致的体现了这样的优势，证明了深度学习挖掘出了有效特征；与 MLP 比较，可以发现，图卷积操作也对电力网络的特征的提取有明显的优势。

| Model | kkd | ryd | ysl | acc |
|---|---|---|---|---|
| Baseline | 98.19 | 65.17 | 72.13 | 81.65 |
| MLP | 98.157 | 59.92 | 74.17 | 84.33 |
| SVM | 98.19 | 83.66 | 36.61 | 46.77 |
| GraphModel | **98.09** | **54.66** | **77.18** | **87.32** |
| GraphPool | 98.08 | 69.03 | 66.32 | 76.44 |
| DeepCNN5 | 98.03 | 73.49 | 61.00 | 71.13 |

c) 展示交叉验证，有 graph 效果明显好；拓扑特征的必要性

为了考察各部分特征（全局特征、局部特征、拓扑特征）对于深度模型性能的影响已经拓扑特征的必要性，我们通过减少部分输入的特征重新训练深度网络，当缺少某部分特征时，该部分特征对应的结构也相应地做删减，来实现交叉验证，最后比较验证集上各指标的性能，如下表所示。当缺少局部特征后，压缩率和准确率均有显著下降，冗余率则显著上升，说明虽然有了拓扑特征，局部特征至关重要。当缺少全局特征，性能也有较大下降；当缺少拓扑特征，性能也有明显下降，但没有上述两种特征影响大。综上所述，为了达到最好的性能，加入拓扑特征可以进一步提高判稳网络的性能，同时拓扑特征并不需要做额外处理，通过图卷积可以较好的抽取出隐含的拓扑特征。

| | kkd | ryd | ysl | acc |
|---|---|---|---|---|
| 全局+局部+邻接矩阵 | 98.09 | 54.66 | 77.18 | 87.32 |
| 缺少全局 feature | 96.38 | 72.27 | 63.35 | 73.13 |
| 缺少局部 feature | 95.72 | 86.13 | 27.19 | 36.834 |
| 缺少 graph 信息 | 97.96 | 64.33 | 71.03 | 81.15 |

a) 展示全局特征效果好

通过对比发现，全局特征的加入

# 误差分析和改进策略：

文献[]中提出可将前述简单方法与含参数方法的预测失稳样本合并，能够提高判稳效果，该方法是最简单的动态故障集生成策略，其优势在于不受样本质量和偏差的影响，劣势在于不能反映系统变化的本质，对于新出现的失稳故障缺乏判断力。其优缺点恰好与本文的扩展边界 SVM 方法互补。具体的结果如下：

| 日期 | kkd | Ryd | Ysl | acc |
|---|---|---|---|---|
| GraphModel-0.0021<br>GraphModel-0.0012 | 98.351<br>**98.62** | 40.26<br>**43.90** | 84.23<br>**83.16** | 93.492<br>**92.47** |
| Baseline | 98.32 | 43.68 | 84.32 | 93.00 |
| Combined（集成模型） | 98.62 | 67.37 | 71.04 | 80.36 |

将基准模型 1 与深度图卷积网络的预测结果进行合并后得到的结果中，可靠性增强了，而相应的其他指标则均有下降，为了达到具有可比性的结果，我们重新设置深度图卷积网络的阈值到 0.0012， 使得深度图卷积网络的阈值和集成模型的可靠性均为 98.62，可以看到，调整后的深度图卷积模型在其余三个指标上，均一致性地好于集成模型。所以本次集成实验在与本文算法结合中并没有明显改进效果，分析这是由于前述简单方法主要考虑了电网在时间上的周期性和连续性，前一天的失稳故障在后一天更可能成为失稳故障。而本文算法中由于加入了节点向量化编码，导致失稳故障的节点能够在节点向量化中得到表征。神经网络可由此学习到相关节点的周期化特性。从实际测试结果中也可看出，缺少了节点向量化的算法模型效果下降到 kkd 95.72， 而完整模型达到 98.09，

# 结论：

本项目针对电网故障延时切除的稳定性建模问题进行了研究，主要的工作包括：

（1）对深度学习研究进展的调研。

首先，确定了电网研究可以参考借鉴数据驱动的机器学习方法的潜在可能。深度学习是机器学习领域近两年最重要的发展，本项目立足于深度学习研究首先调研了深度学习在卷积网络，循环网络，图卷积等重要方面的进展。之后调研了在图像，语音，语言等典型机器学习领域深度学习的前沿应用。

通过上述工作，并且结合相关深度学习在电力领域的少数应用报道，借鉴并确立了本方案的技术路线。确定利用以图卷积为代表的深度学习网络解决电力系统中稳定性建模问题。

（2）通过以深度学习实现电网特征的自动挖掘

利用图卷积为代表的深度学习系统，挖掘电网特征。本项目首先分析结合了电网的物理特性和结构特点，电网中的拓扑结构适合用图卷积进行建模处理，同时电网的局部和全局特征对稳定判别都有作用，本项目通过复合多元的神经网络进行分别建模。最终通过训练得到了具有明确区分意义的电网特征。

（3）对稳定性的快速判断。

利用东北电网的实际运行数据，本项目在深度图卷积网络的基础上进行了实际数据测试。结果显著显示，相比于基于规则的方法和传统 svm 等机器学习方法，本项目提出的深度图卷积网络方法能够最大限度提高预测准确率和召回率，具有潜在的应用价值。

## 参考文献